\documentclass[letterpaper, 10 pt, journal, twoside]{IEEEtran}
\usepackage{graphics} 
\usepackage{epsfig} 
\usepackage{mathptmx} 
\usepackage{times} 
\usepackage{amsmath} 
\usepackage{amssymb}  
\usepackage{gensymb}
\usepackage{multirow}
\usepackage{hhline}
\usepackage{algorithm}  
\usepackage{algpseudocode}  
\usepackage{cite}
\usepackage{subfigure}       
\usepackage{graphicx}
\usepackage{color}

\ifCLASSINFOpdf
\else
\fi
\begin{document}
%

\title{Semantic Histogram Based Graph Matching for Real-Time Multi-Robot Global Localization in Large Scale Environment
}

\author{Xiyue Guo$^{1}$, Junjie Hu$^{1}$, Junfeng Chen$^{1}$, Fuqin Deng$^{1}$, and Tin Lun Lam$^{2,1,\dagger}$
\thanks{Manuscript received: October 15, 2020; Revised: January 7, 2021; Accepted: February 4, 2021.}
\thanks{This paper was recommended for publication by Editor Javier Civera upon evaluation of the Associate Editor and Reviewers’ comments.}
\thanks{This work was supported by the National Key R$\&$D Program of China (2020YFB1313300) and funding 2019-INT008 from the Shenzhen Institute of Artificial Intelligence and Robotics for Society.}
\thanks{$^{1}$The Shenzhen Institute of Artificial Intelligence and Robotics for Society.}%
\thanks{$^{2}$The Chinese University of Hong Kong, Shenzhen.}%
\thanks{$^{\dagger}$Corresponding author is Tin Lun Lam
        {\tt\small tllam@cuhk.edu.cn}
        }%
\thanks{Digital Object Identifier (DOI): see top of this page.}
}

\markboth{IEEE Robotics and Automation Letters. Preprint Version. Accepted February, 2021}
{Guo \MakeLowercase{\textit{et al.}}: Semantic Histogram Based Graph Matching for Real-Time Multi-Robot Global Localization} 

%



\maketitle

\begin{abstract}
The core problem of visual multi-robot simultaneous localization and mapping (MR-SLAM) is how to efficiently and accurately perform multi-robot global localization (MR-GL). The difficulties are two-fold. The first is the difficulty of global localization for significant viewpoint difference.
Appearance-based localization methods tend to fail under large viewpoint changes. Recently, semantic graphs have been utilized to overcome the viewpoint variation problem. However, the methods are highly time-consuming, especially in large-scale environments. This leads to the second difficulty, which is how to perform real-time global localization. In this paper, we propose a semantic histogram based graph matching method that is robust to viewpoint variation and can achieve real-time global localization. Based on that, we develop a system that can accurately and efficiently perform MR-GL for both homogeneous and heterogeneous robots. The experimental results show that our approach is about 30 times faster than Random Walk based semantic descriptors. Moreover, it achieves an accuracy of 95\% for global localization, while the accuracy of the state-of-the-art method is 85\%.
\end{abstract}

\section{INTRODUCTION}
\begin{figure}[t!]
\centering
\includegraphics[width=3in]{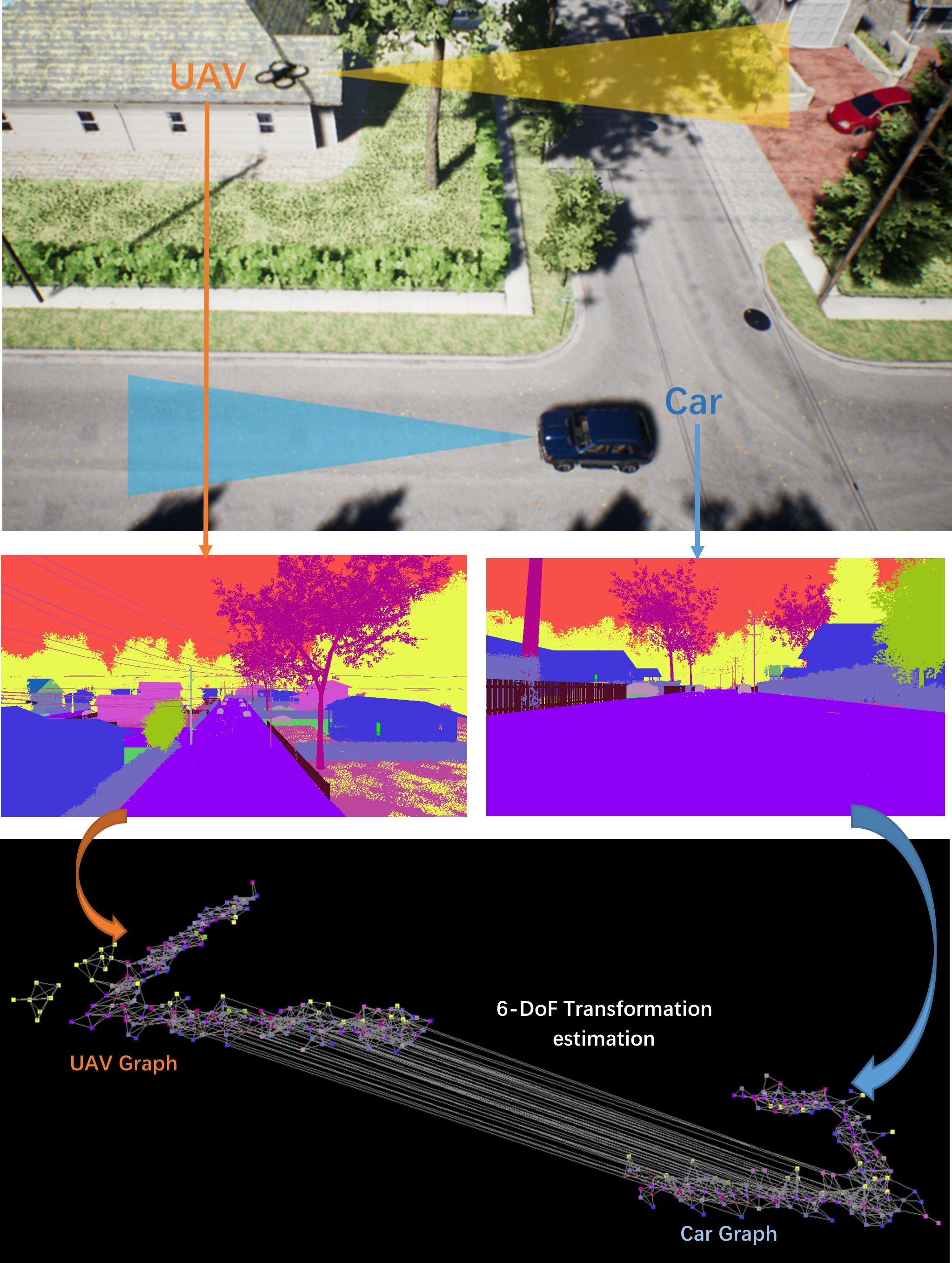}
\centering
\vspace{-2mm}
\caption{An example of our semantic based graph matching method. The method is used for global localization in a large scale environment. The viewpoint between two robots (UAV and Car) is extremely large. 
We utilize semantic maps to build semantic graphs for two robots. Then, the transformation matrix between them can be simply estimated.}
\label{fig:biaoti}
\vspace{-5mm}
\end{figure}

Vision based single-robot simultaneous localization and mapping (SR-SLAM) have gained significant progress over the past decades. However, it has fundamental limitations of mapping speed, mission range, localization accuracy, etc. Thus, it usually performs poorly for large scale environments. To overcome these problems, multi-robot SLAM (MR-SLAM) maps large scale unknown environments by exploiting several collaborating robots\cite{survey,carlone2010rao}.
Although it has clear advantages against SR-SLAM with multi-robot cooperation, 
the preliminary difficulty is that we need the multi-robot global localization to satisfy the requirement of real-world deployment.

There are mainly two difficulties towards achieving this goal. First, an urgent problem is the accurate global localization for the large viewpoint difference between individual robots\cite{X-view}.
It's known that MR-SLAM is applicable for large scale environments. The viewpoint differences between robots are very large. The difference is more significant for heterogeneous robot systems.
An example is given in Fig.~\ref{fig:biaoti}, the images captured by a vehicle show a clear viewpoint difference from those captured by a drone.
Second, the global localization has to be computationally efficient\cite{lee2012survey}; otherwise, the MR-SLAM will collapse. 

In previous studies, traditional appearance-based methods such as the Bag-of-Word (BoW)\cite{bow,FAB} are the most widely used localization methods for MR-SLAM\cite{CCM,dis-SLAM}. However, Under the large viewpoint changes, local image features (e.g. SIFT\cite{SIFT}, SURF\cite{SURF}, ORB\cite{ORB-detector}, FAST\cite{FAST}) will change significantly, this cause appearance-based methods to fail.
As semantic information is invariant to the viewpoint changes, recently, several methods \cite{X-view,indoor1, indoor2} proposed to use semantics for global localization. In these works, semantic based graphs are first built for different viewpoints, then the different view's graphs are matched by utilizing the semantic information. 
These methods demonstrate better performance compared with appearance-based methods for large viewpoint changes.

In large scale environments, there will be many mismatches if we directly perform graph matching with only the semantic label of each node. Hence, for each node, the descriptor should be extracted to contain the surrounding information. In the previous methods\cite{X-view, indoor1}, the Random Walk based descriptors are utilized for graph matching. On the other hand, no matter how large the graphs are, the graph matching needs to be processed in real-time. 
This requirement eliminates those Random Walk based descriptors  \cite{X-view,indoor1}, which are highly time-consuming.

\begin{figure}[t!]
\centering
\includegraphics[width=3.4in]{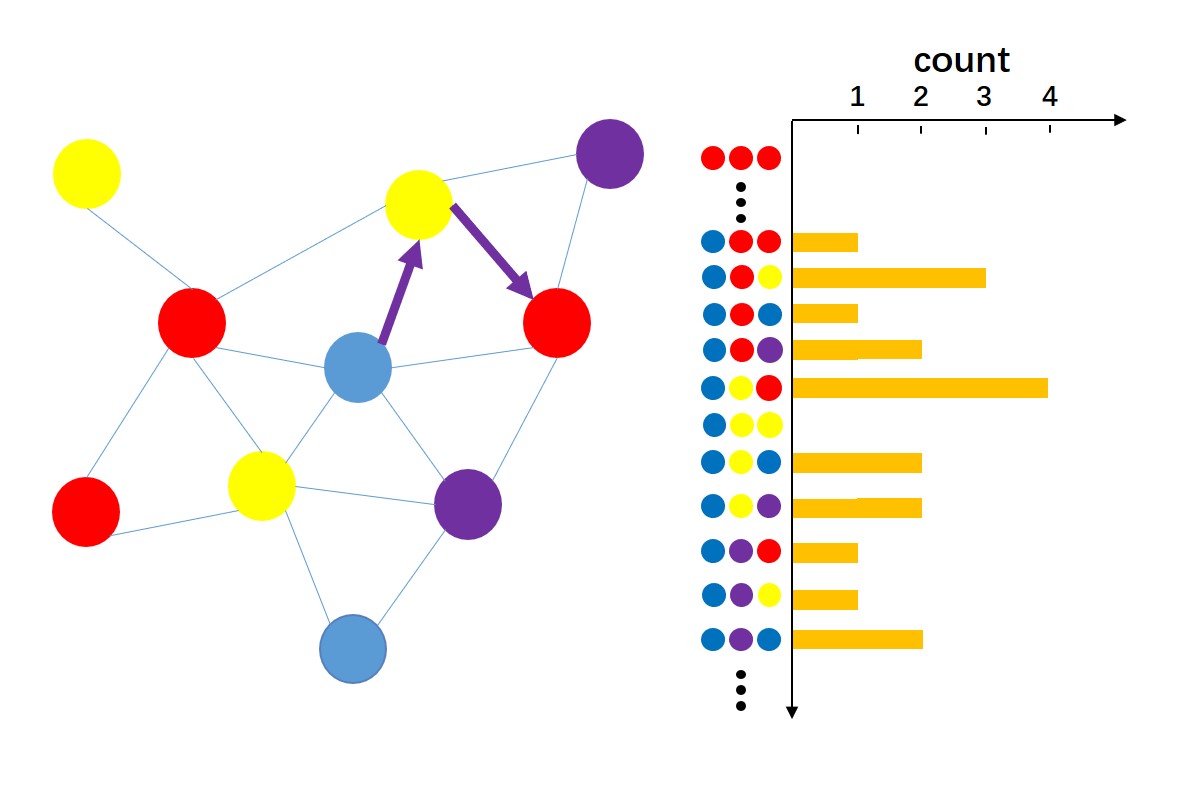}
\vspace{-8mm}
\centering
\caption{An illustration of the semantic histogram based descriptor. Left is the semantic graph. The searched path is started from the start point (blue). The path information is recorded as a prearranged histogram on the right. The similarity score between two descriptors can be obtained through the normalized dot-product.}
\label{fig:descriptor}
\vspace{-5mm}
\end{figure}

In this paper, we propose a more accurate and computationally efficient method. Our method is based on the semantic based graph matching. A novel semantic histogram based descriptor is proposed to enable a real-time matching under the large viewpoint changes. The descriptor stores the surrounding paths' information in the form of a prearranged histogram. Fig.~\ref{fig:descriptor} shows the illustration of the descriptor. Based on the new descriptor, we further develop a semantic graph-based global localization system to merge maps for MR-SLAM. Our method is fairly tested on three datasets, including two synthesized datasets and a  publicly available real-world dataset. As a result, 
we show through the experiments that

\begin{itemize}

\item 
Our method outperforms both appearance and semantic based methods by a large margin. 
It performs stably and accurately for large viewpoint differences in large scale environments.

\item
Our method is much faster than the state-of-the-art semantic based method \cite{X-view}.  It yields satisfactory performance for both homogeneous and heterogeneous robot systems.

\item Our method demonstrates good performance for map fusion in large scale real-world KITTI dataset in which we only take RGB images as input, the depth maps and semantic maps are predicted by deep convolutional networks.

\end{itemize}

\section{RELATED WORK}
\subsection{Appearance-based Approaches}


Appearance-based localization methods such as Bag-of-Words (Bow) use global or local visual features to find the association of images \cite{SIFT,SURF,ORB-detector,FAST}. One representative work is FAB-MAP\cite{FAB}.
 These methods work well under the small viewpoint difference. However, when the viewpoint difference is large, the localization systems become less reliable. 

Recently, convolution neural networks (CNNs) have been employed to overcome the viewpoint change problem. In \cite{ConvNet,DeepLearning,NetVLAD}, the viewpoint invariant landmarks are generated by CNNs. However, these landmarks are not reliable when the viewpoint change becomes significant (e.g., opposite direction, viewpoint changes of heterogeneous multi-robots, such as a car and a UAV).  Several approaches are proposed to overcome the opposite viewpoint problem. LoST\cite{lost} uses semantics and appearance information to recognize the places in the opposite viewpoint. In \cite{loolnoDeeper}, the place recognition in the opposite viewpoint is handled by matching a sequence of depth-filtered keypoints with a single-query image's keypoints. However, these approaches are only focused on place recognition, the localization is not considered. As the normal approaches such as ORB-SLAM\cite{ORB-SLAM3} does not work well for large viewpoint difference, thus the pose estimation after place recognition has to be carefully handled \cite{angelina2018pointnetvlad}.

\subsection{Graph-based Approaches}
Graph-based methods formulate the global localization problem as a graph registration problem. The associations between different graphs are found by extracting the correspondences between nodes across the graphs. Then, the relative pose between graphs can be calculated. 
In \cite{graph1,graph2}, each node is labeled by local features based visual word. 
However, as mentioned before, the appearance features are not reliable when the viewpoint changes are significant. 

Recently, several methods employ semantics to generate labels \cite{indoor2,graphseg1}. In \cite{graphseg1}, brute force is used to match two graphs based on their semantic labels. Unfortunately, this method can only work for simple and limited environments where the number of objects and the environments' scale are small. The same problem also exists in\cite{indoor2}, in which semantic graphs are matched with the Hungarian algorithm. In large scale environments, the same objects frequently appear in multiple places, such as cars and buildings in city blocks.
Such environments will lead these methods to fail.
To enable more accurate semantic based matching, Random Walk method is used to generate descriptor for each semantic node \cite{X-view,indoor1}.
However, when the matching graphs are large, the computational complexity becomes extremely high. This will largely hinder the deployment of these methods to real-world applications such as multi-robot SLAM.
In this paper, we present a semantic histogram based descriptor, which enables the graph matching to be performed in real-time and it is even more accurate compared to Random Walk based methods.



\begin{figure*}[t!]
\centering
\includegraphics[width=6in]{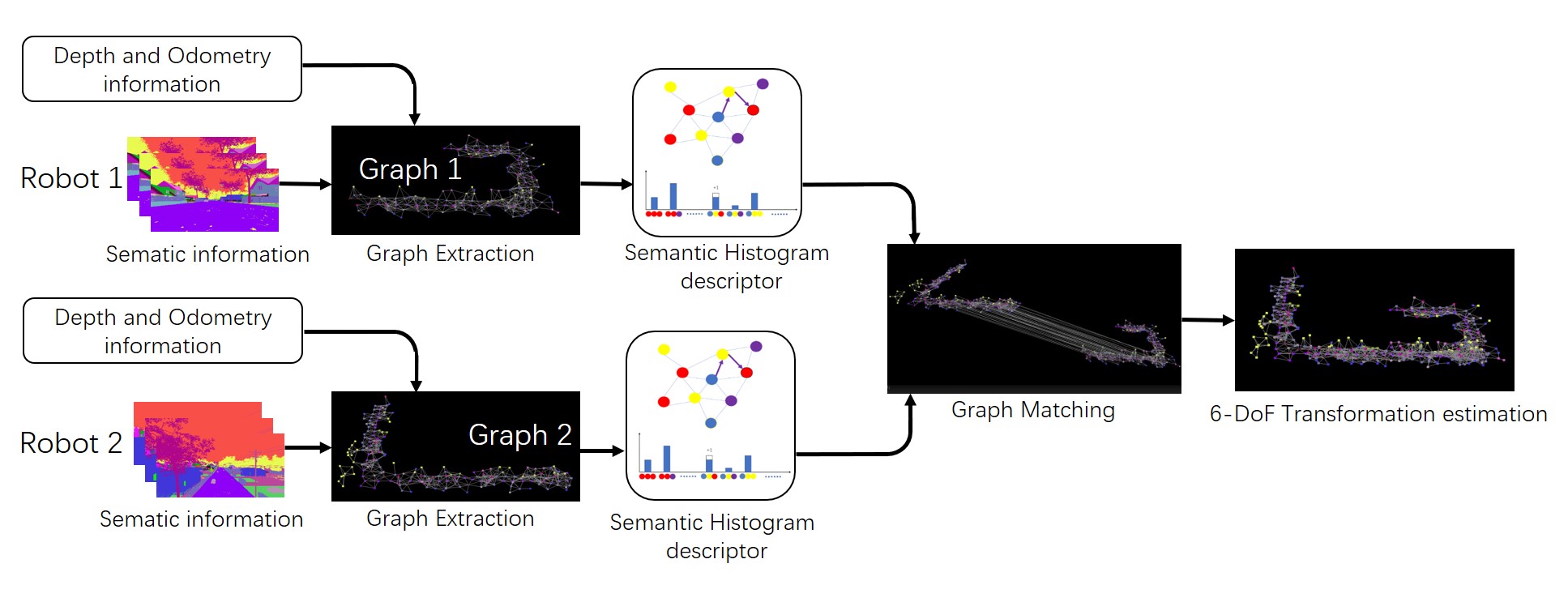}
\centering
\vspace{-6mm}
\caption{The diagram of our global localization system. The system takes semantic maps, depth maps, and odometries as inputs. The 3D semantic graph for each robot is first built from the inputs. Then, the descriptor of each node is extracted with the semantic histogram method. Next, the two graphs are matched by comparing the descriptors across the graphs. Finally, the matched correspondences are used to estimate the 6-DoF transformation between the coordinate systems of two robots.}
\label{fig:overview}
\end{figure*}
\section{Semantic Graph-based Global localization}
In this section, we present our semantic histogram based graph matching system for global localization. Our framework is partially inspired by X-view\cite{X-view}. Firstly, given two odometries, related depth maps, and semantic maps. We first generate semantic graphs. Then, the semantic histogram based descriptors are extracted. The two graphs are matched with the extracted descriptors. Finally, the 6-DoF transformation matrix is calculated. 
The framework of our global localization system is shown in Fig.~\ref{fig:overview}. 

\subsection{Graph Extraction}
Similar to \cite{X-view}, to build the graph, we need to extract nodes from images. Towards this end, we employ the seed filling method\cite{seedfilling} to segment objects from images.
To avoid the failed segmentation between two neighboring objects with the same semantics, we use the 3D coordinates of pixels during the segmentation process. Then, the 3D geometry center of each object is extracted as a node. 
It's noted that nodes with the same semantic label should be merged if they are highly close to each other. 
Therefore, each node contains two types of information: 1), 3D coordinates value of the node; 2), The semantic label. 
The undirected edges between the nodes are then formed if the distances between nodes are smaller than a set threshold (connectivity threshold). Finally, nodes and edges together form the semantic based graph.

\renewcommand{\algorithmicrequire}{\textbf{Input:}}  
\renewcommand{\algorithmicensure}{\textbf{Output:}}  
\begin{algorithm}[t!]  
  \caption{Descriptor Extraction}  
  \label{alg::Descriptor_generation}  
  \begin{algorithmic}[1]  
    \Require  
      $G$: Semantic Graph;
    \Ensure  
      $V$: Histogram of path descriptors for $G$;
    \For{$i$-th node in $G$}
        \State Initialize the histogram vector $V_i$;
        \State Record the node's label $l_i$;
        \For{m in neighbor nodes of i}
            \State Record the first neighbor node's label $l_m$;
            \For{n in neighbor nodes of m}
                \State Record the second neighbor node's label $l_n$;
                \State The Histogram cell $V_i$($l_i$-$l_m$-$l_n$) plus one;
            \EndFor
        \EndFor
       \State Add $V_i$ into $V$;
    \EndFor
 \end{algorithmic}  
\end{algorithm}

\subsection{Semantic Histogram Based Descriptor}
In order to describe each node in the graph, the surrounding information of the node needs to be recorded by extracting the node's descriptor. For the semantic graph, histogram based descriptors are simple and feasible; furthermore, the matching procedure of this type of descriptors is very fast. Intuitively, the simplest histogram based descriptor is the Neighbor Vector descriptor \cite{graph2}. It describes the node by counting all the neighbor nodes' labels. However, due to the lack of topology information, the matching performance of the neighborhood vector is low \footnote{We will confirm this through our experiments evaluations in Sec.~\ref{AirSim-experiments} and Sec.~\ref{real-world-experiments}.}.

Therefore, we propose to include more surrounding information for all nodes. To be specific, for each node, the descriptor stores all possible paths that started from it. We set the length of the path as 3. Therefore, each path can be considered as a 3 dimensional vector, recording the three steps' semantic labels. For a single descriptor, all possible paths are counted in the form of the prearranged histogram. Therefore, the topology information of objects and their neighbors are stored in descriptors. 
The illustration of our descriptor is shown in Algorithm~\ref{alg::Descriptor_generation}.
The proposed descriptor enables the graph matching to be much computationally efficient. The time complexity of one descriptor extraction is $O(MN)$, where $M$ and $N$ are the numbers of first-order and second-order neighbors, respectively.

\subsection{Graph Matching}

\begin{algorithm}[t!]  
  \caption{Graph matching and ICP-RANSAC rejection}  
  \label{alg::graph_matching}  
  \begin{algorithmic}[1]  
    \Require  
      $V$, $V'$: descriptor sets of two graphs; $N_i$: iteration number for RANSAC; $M_0$: initial matches set;
    \Ensure  
      $M_1$: final matches set;
    \State Initialize $M_0$;
    \For{i in $V$}
        \For{j in $V'$}
            \State scores = Score($V_i$ , $V'_j$);
            \If{scores $>$ score threshold $T_s$};
                \State Add the Correspondence $C_{ij}$ to $M_0$;
            \EndIf
        \EndFor
    \EndFor
    \State Initialize $M_1$;
    \State Initialize the Maximum Inlier number $A^*$;
    \State let $A^* = 0$;
    \For{o = 1 to $N_i$}
        \State Select 4 correspondences $M_{four}$ Randomly;
        \State $R_o$, $t_o$ = ICP($M_{four}$);
        \For{k in Matches set $M_0$}
            \State Obtain the correspondence $C_{k}$;
            \State Error = Evaluation($C_{k}$, $R_o$, $t_o$);
            \If{Error$<$ Threshold $T_R$}
                \State Add $C_k$ to the Inlier set $M_o$;
            \EndIf
        \EndFor
            \State Inlier number $A$ = Count($M_o$);
            \If{$A>A^*$}
                \State $M_1 = M_o$;
                \State $A^* = A$;
            \EndIf
    \EndFor
\end{algorithmic}  
\end{algorithm}

Similar to image matching, the descriptors of nodes are compared across the graphs by computing the similarity scores. 
In the matching process, only the nodes that have the same labels will be compared. The similarity score is obtained by taking the normalized dot-product between two descriptors. It is formulated as follows:
\begin{equation}\text { Score(A, B) }=\frac{\sum_{d=1}^{n_d} A_{d} \times B_{d}}{\sqrt{\sum_{d=1}^{n_d}\left(A_{d}\right)^{2}} \times \sqrt{\sum_{d=1}^{n_d}\left(B_{d}\right)^{2}}}\end{equation}
where A and B denote descriptors of nodes from two graphs. $n_d$ is the descriptor dimension, which is equal to the cubic of the label number $n_l$. The time complexity of one pair nodes' matching is $O(n_d)$, the size of $n_d$ is typically on the order of hundreds.  

The similarity score between the two nodes are between 0 and 1, the higher score means higher similarity. The correspondences whose similarity scores are higher than the threshold $T_s$ are stored as the matching candidates. 
Note that these matching candidates still contain many incorrect matches. In graph-based localization, since the transformation between two graphs is rigid. The transformation values between the correct pairs of correspondences should all be similar. This condition can be utilized to reject the incorrect matches (outliers).  In order to guarantee these consistency of the correspondences, the ICP-RANSAC algorithm\cite{ICP,RANSAC} is used to reject the outliers.
Finally, the remained inlier correspondences are kept for the pose estimation method. In addition, the rotation matrix $R$ and the translation vector $t$, which are obtained from the ICP-RANSAC method are stored as the initial value of the pose estimation method. 
The illustration of the graph matching is shown in Algorithm~\ref{alg::graph_matching}. 

\subsection{Pose Estimation}
In this step, the final transformation matrix is computed with ICP algorithm. In the method, the inlier correspondences obtained by RANSAC method are used for registration. Hence, the Rotation matrix $R$ and translation vector $t$ is obtained by minimizing the sum of squared error:
\begin{equation}E(R, t)=\frac{1}{N_{p}} \sum_{k=1}^{N_{p}}W_k\left\|q_{k}-R p_{k}-t\right\|^{2}\end{equation}
The $N_p$ is the correspondences number after RANSAC rejection. $q_k$ and $p_k$ are the correspondent nodes in two graphs. $W_k$ is the weight element, which is related to the corresponding objects' size.

\section{Experiment Results}
To fairly and fully validate the effectiveness of our method, we conduct three experiments on multiple datasets. First, we show the quantitative comparisons between our method and previous approaches on the SYNTHIA dataset \cite{SYNTHIA}. Second, we show the performance of our method for multi-robot global localization, we apply our method to both homogeneous and heterogeneous multi-robot systems. Thirdly, to verify the generability of our method, we conduct another experiment on the real-world KITTI dataset, where we only use RGB images as input. Finally, we investigate the effect of different parameter settings and input qualities.  
All the experiments are computed on an Intel Core i7-8565U CPU @ 1.80GHz.

\begin{figure}[t!]
\centering
\includegraphics[width=3.4in]{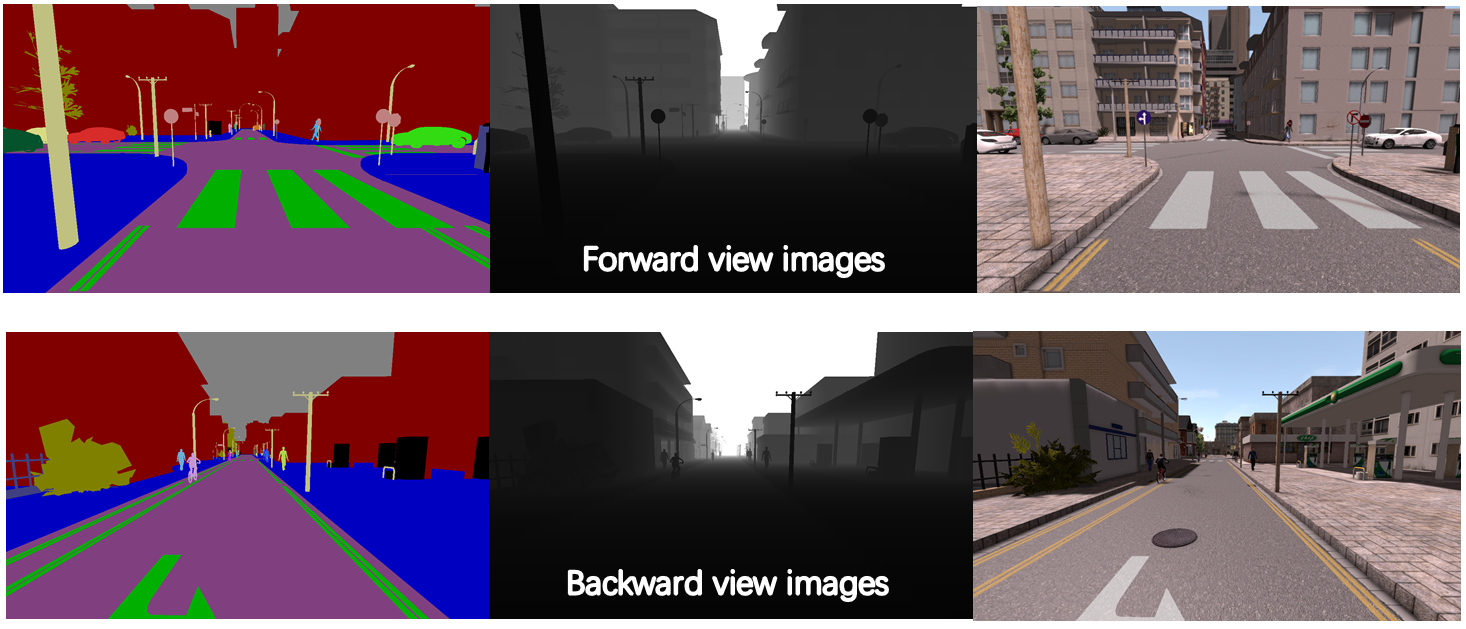}
\centering
\caption{Samples of SYNTHIA dataset. images in top row are the forward view images, including semantics, depths, and RGB images. The images in bottom row are the backward view images collected at the same time.}
\label{fig:SYNTHIA}
\end{figure}

\subsection{Performance Comparison}
\subsubsection{Dataset and Implementation Details}

The SYNTHIA dataset collects data from sensors mounted on a simulated car in a dynamic urban environment. In our experiment, we use sequence 04-spring as our test sequence. It contains three types of data, including RGB images, depth maps, and semantic maps. All these images have four camera directions, i.e., forward, backward, leftward, and rightward. In order to simulate the viewpoint variation, data of the forward view is picked to be associated with backward view's data, as seen in Fig.~\ref{fig:SYNTHIA}. The travel distance of the dataset is 950 meters.  In the experiment, graphs generated by 30 backward view's frames are matched with the global forward view's graph.

Several previous methods are taken as baseline methods. The first is
X-view\cite{X-view}, which is the state-of-the-art of semantic graph-based global localization. The second is an appearance-based Bag-of-Words (BoW) method that is built on the DBoW3 library\cite{bow}. The third method is a CNN-based method called NetVLAD \cite{NetVLAD}.
In order to have a fair comparison, the experiment setting is made completely the same as X-view.
Since there is no available open-source code of X-view, we directly use the results in its paper \cite{X-view}. 
Similarly, the results of NetVLAD are also taken from \cite{X-view}. In addition, the performance of another fast descriptor called Neighbor Vector\cite{graph2} is also presented in the plot.
\begin{figure}[t!]
\centering
\includegraphics[width=3.2in]{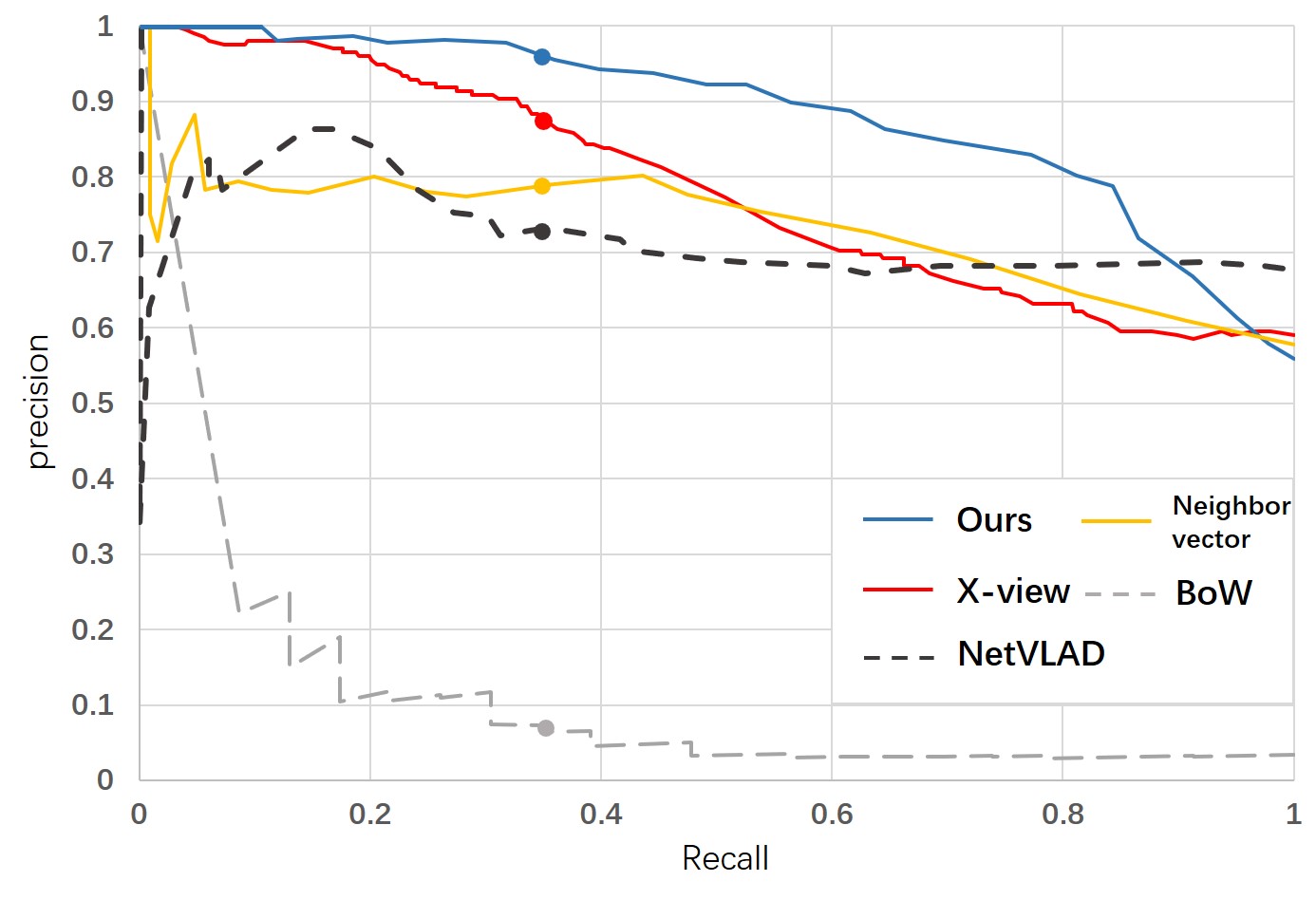}
\centering
\vspace{-4mm}
\caption{Precision-Recall curve of different global localization methods. The operation points are shown as dots.}
\label{fig:PR}
\end{figure}

\subsubsection{Experiment Results}
In the experiment, the performance of the localization is represented by the Precision-Recall curve (PR-curve). The precision depends on the localization threshold $T_p = 20m$. The localization is set to true if the translation error is lower than 20 meters. The recall is controlled by the variable threshold of $T_r$. If the inlier number obtained from the ICP-RANSAC method is higher than $T_r$, then this localization gets a positive vote. 

The results are shown in Fig.~\ref{fig:PR}, it's clear that
the semantic graph-based method is more accurate than the appearance-based method when the viewpoint change is significant. Moreover, our method shows a clear advantage against X-view as it outperforms it by a large margin. To be consistent with X-view, we use the same operation point (recall is 0.35) to compare the success rate of global localization; as a result, our method gained $95\%$ success rate while
the success rate of BoW, X-view, NetVLAD, and Neighbor Vector are $8\%$, $85\%$, $73\%$,and $79\%$ respectively.
In addition, the time complexity for every component of our method is shown in Table \ref{time_cost}.

\subsection{Global Localization for Multi-robots} 
\label{AirSim-experiments}
\subsubsection{Dataset and Implementation Details}
We consider yet another problem that is the global localization for multiple large scale odometries generated by multiple robots. This is a key step for map fusion of multi-robot SLAM. We evaluate the performance of our method for both heterogeneous and homogeneous robot systems.

\begin{figure}[t!]
\centering
\includegraphics[width=3.4in]{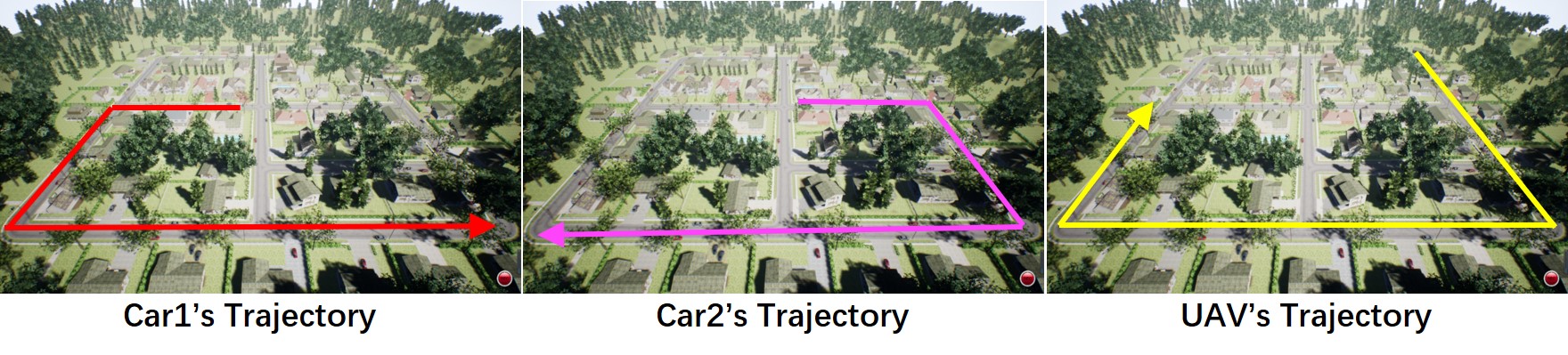}
\centering
\vspace{-4mm}
\caption{The illustration of three simulated trajectories generated from AirSim. We use them to evaluate the performance of our global localization method for both homogeneous and heterogeneous robot systems.}
\label{fig:air2}
\end{figure}

As there's no publicly available dataset for this purpose, we manually create a dataset \footnote{The dataset will be made publicly avaliable.} from the Neighborhood (an urban block) of AirSim \cite{AirSim}.
We generate two Car's trajectories, and one UAV trajectory. Therefore, there are three trajectories in total. Same as the SYNTHIA, the dataset contains three types of data, i.e. RGB images, depth maps, and semantic maps. 
The average travel distance of Car is 420 meters and the travel distance of the UAV is about 600 meters. The viewpoint change between them is extremely large.
The detailed illustrations of these trajectories are shown in figure \ref{fig:air2}. It's seen that the different trajectories contain long overlapping parts (over 200 meters); meanwhile, they have their own non-overlapping parts. 

There are three descriptors compared in the experiment. The first one is the Random Walk, which is used in X-view. The second one is Neighbor Vector \cite{graph2}. The final one is our semantic histogram based descriptor. In the experiment, we use the whole graphs to perform localization. The matching performance is evaluated by using good matches number and good matches rate.  We halve the threshold of correct graph matching (20 meters) as the criteria to identify good matches, i.e. 10 meters. The good match rate is the percentage of the correct matching correspondences represented in the total correspondences. 

\begin{table}[t!]
\caption{Time-consuming for each component of our method.}
\vspace{-4mm}
\label{time_cost}
\begin{center}
\begin{tabular}{c|c}
\hline
Module                & Time(ms/frame) \\ \hline
Graph Extraction      & 114.23 $\pm$ 4.53\\
Descriptor Extraction & 0.68  $\pm$ 0.03     \\
Graph Matching              & 1.65  $\pm$ 0.49   \\
Pose Estimation          & 0.63  $\pm$ 0.14   \\ \hline
Total          & 117.19  $\pm$ 5.19   \\ \hline
\end{tabular}
\end{center}
\vspace{-5mm}
\end{table}

\begin{table*}[]
\begin{center}
\caption{The quantitative comparisons of different descriptors for global localization of multi-robot systems on AirSim.}
\label{matching}
\begin{tabular}{ccc|ccc|ccc}
\hline
Robot Type &
  \begin{tabular}[c]{@{}c@{}}Matching \\ Graph \\ Size\end{tabular} &
  \begin{tabular}[c]{@{}c@{}}Descriptor\\ Type\end{tabular} &
  \begin{tabular}[c]{@{}c@{}}Matching\\ Time\\ (sec)\end{tabular} &
  \begin{tabular}[c]{@{}c@{}}Good\\ Matches\end{tabular} &
  \begin{tabular}[c]{@{}c@{}}Good\\ Matches\\ Rate(\%)\end{tabular} &
  \begin{tabular}[c]{@{}c@{}}Processing\\ Time\\ (sec)\end{tabular} &
  \begin{tabular}[c]{@{}c@{}}Translation\\ Error\\ (m)\end{tabular} &
  \begin{tabular}[c]{@{}c@{}}Rotation\\ Error\\ (degree)\end{tabular}\\ \hline
\multirow{3}{*}{\begin{tabular}[c]{@{}c@{}}Car1\\ and\\ Car2\end{tabular}} &
  \multirow{3}{*}{\begin{tabular}[c]{@{}c@{}}317 points\\ and\\ 328 points\end{tabular}} &
  Random Walk &
  4.155 &
  125 &
  40.0 &
  4.304 &
  $3.44 \pm 1.39$ &
  0.92 $\pm$ 0.49\\
 &
   &
  Neighbor Vector &
  \textbf{0.013} &
  120 &
  37.8 &
  \textbf{0.057} &
  $4.55 \pm 1.32$ &
  0.48 $\pm$ 0.31\\
 &
   &
  Ours &
  0.132 &
  \textbf{152} &
  \textbf{49.1} &
  0.184 &
  \textbf{3.12 $\pm$ 0.76} &
  \textbf{0.30 $\pm$ 0.29}\\ \hline
\multirow{3}{*}{\begin{tabular}[c]{@{}c@{}}Car1\\ and\\ UAV\end{tabular}} &
  \multirow{3}{*}{\begin{tabular}[c]{@{}c@{}}317 points\\ and\\ 486 points\end{tabular}} &
  Random Walk &
  6.637 &
  136 &
  43.6 &
  6.859 &
  $2.23 \pm 0.88$ &
  2.62 $\pm$ 0.46\\
 &
   &
  Neighbor Vector &
  \textbf{0.021} &
  100 &
  31.7 &
  \textbf{0.089} &
  $3.23 \pm 1.02$ &
  3.25 $\pm$ 0.54\\
 &
   &
  Ours &
  0.195 &
  \textbf{142} &
  \textbf{45.1} &
  0.248 &
  \textbf{2.12 $\pm$ 0.47} &
  \textbf{2.61 $\pm$ 0.25}\\ \hline
\end{tabular}

\end{center}
\end{table*}

\subsubsection{Experimental Results}

The quantitative comparisons are shown in Table~\ref{matching}, where the results are average for 100 times experiments.
It's clear that our method achieves the lowest translation and rotation error for global localization. The average translation and rotation error of Random Walk, Neighbour Vector, and our semantic histogram method are 2.83, 3.89, and 2.62 meters; 2.23, 1.86, and 1.45 degrees, respectively. The lowest translation and rotation error of our method is attributable to the highest good matches rate. It is also observed that although the translation errors are similar, the rotation errors of heterogeneous system are much higher than homogeneous system.

By considering the time complexity, the Neighbor Vector descriptor has the lowest time complexity. However, due to the less surrounding information, the matching performance of the Neighbor Vector is the worst. The time complexity of our descriptor is higher than Neighbor Vector but much lower than the Random Walk. Hence, by considering the trade-off between matching performance and time complexity, our descriptor is the best for semantics graph matching. Overall, our method demonstrates the best performance for the large viewpoint difference between both heterogeneous and homogeneous robot systems.

\begin{figure}[t!]
\centering
\subfigure[The trajectories of KITTI 08 dataset]{
\centering
\includegraphics[width=1.5in]{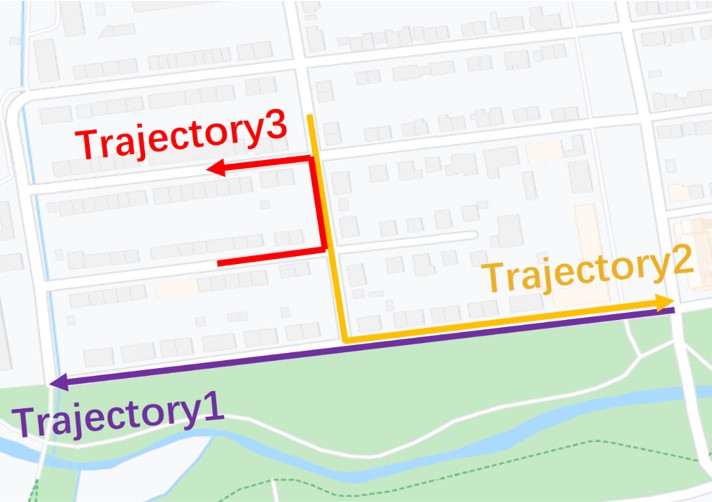}
}
\subfigure[The successful  multi-robots map fusion]{
\centering
\includegraphics[width=1.5in]{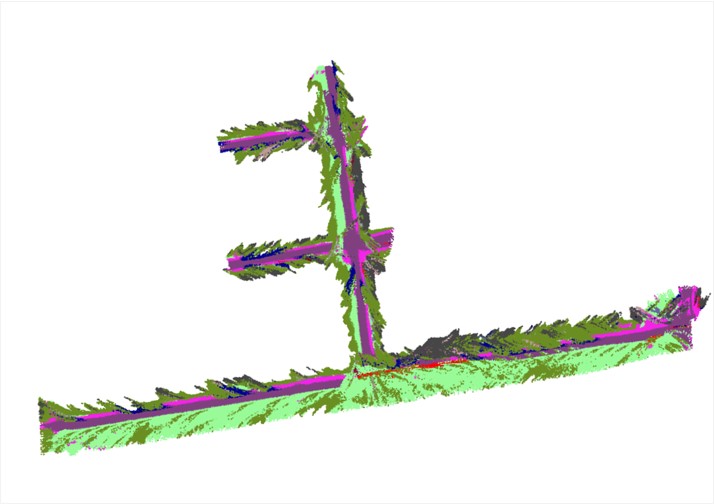}
}
\vspace{-2mm}
 \caption{Trajectories and reconstructed maps of sequence 08 from KITTI dataset.
a) shows the trajectories where each line denotes a trajectory. b) shows the maps reconstructed and merged from these three trajectories.}
\label{fig:RealWorld}
\end{figure}

\subsection{Generability on  Real-World Scenarios}

\label{real-world-experiments}
\subsubsection{Dataset and Implementation Details}
To evaluate the generability of our method in real-world environments, we conduct yet another experiment on the KITTI dataset \cite{KITTI}. To be specific, we evaluate our method on three sequences, sequence 02, 08, and 19. 
In the experiment, three
trajectories are split from the sequence 08. The illustration of
the three trajectories are shown in Fig.~\ref{fig:RealWorld} (a). For simplicity, we use 08A and 08B to represent the alignment between Trajectories 1 and Trajectory 2, Trajectory 2 and Trajectory3. 
For sequence 02 and 19, there are two trajectories which share overlaps in opposite directions.
The total travel length of sequence 02, 08, and 19 are 260 meters, 850 meters and 1000 meters respectively. In addition, the overlap of sequence 02, 08A, 08B, and 19 are 30 meters, 200 meters, 50 meters, and 300 meters respectively. Same as the AirSim, we use the whole graphs to perform the localization.

As the dataset only contains RGB images, there are no ground truths of depth maps and semantic maps. Therefore, we apply current advanced algorithms to estimate depth maps and semantic maps, respectively. The depths are predicted with the method of \cite{depth}, and the semantics are estimated with \cite{segmentation}. For odometry estimation, we use ORB-SLAM3 \cite{ORB-SLAM3}.
For quantitative comparisons, we conduct two types of experiments. The first is the comparison between different descriptors for graph-based global localization as Sec.~\ref{AirSim-experiments}, the second is the comparison against benchmark methods, including BoW\cite{bow} and NetVLAD\cite{NetVLAD}. For these benchmark methods, we simply use the distance between the best matching frames as the translation errors of global localization.
%


\begin{table*}[t!]
\caption{The translation error of global localization on the KITTI dataset (in meters)}
\vspace{-5mm}
\label{KITTI_error}
\begin{center}
\begin{tabular}{c|cccc}
\hline
 & Sequence 02& Sequence 08A & Sequence 08B & Sequence 19 \\ \hline

Neighbor Vector & 14.42$\pm$20.02  & 4.59$\pm$0.63  & 18.42$\pm$4.00 & 15.18$\pm$11.45        \\
Random Walk     & 76.61$\pm$36.42  & 4.83$\pm$0.68  & 25.55$\pm$8.72& 14.63$\pm$13.35        \\ \hline
BoW  & 55.20 $\pm$ 42.01   & 74.12 $\pm$ 51.14  & 32.16 $\pm$ 20.79  & 108.83 $\pm$ 54.05               \\
NetVLAD\cite{NetVLAD}   & 28.21 $\pm$ 19.35 & 35.02$\pm$21.04    & 24.52$\pm$14.41     & 55.11$\pm$20.96            \\ \hline
Ours    & \textbf{8.77$\pm$11.39} & \textbf{4.42$\pm$0.35} & \textbf{7.48$\pm$3.67} & \textbf{8.10$\pm$6.63} \\ \hline
\end{tabular}
\end{center}
\end{table*}

\begin{figure*}[t!]
\vspace{-3mm}
\centering
\subfigure[Connectivity]{
\centering
\includegraphics[width=2.0in]{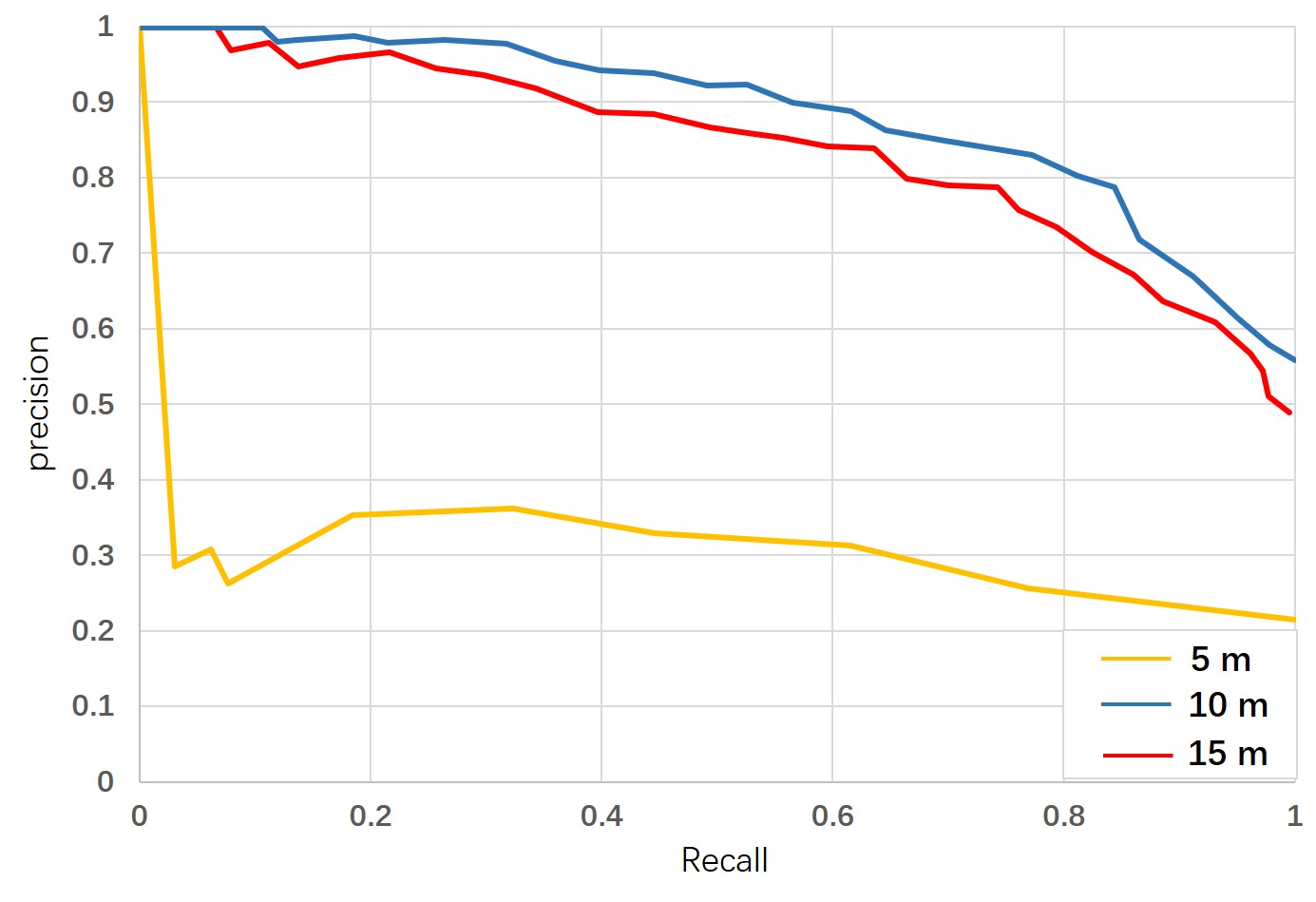}
}
\subfigure[Path dimension of descriptors]{
\centering
\includegraphics[width=2.0in]{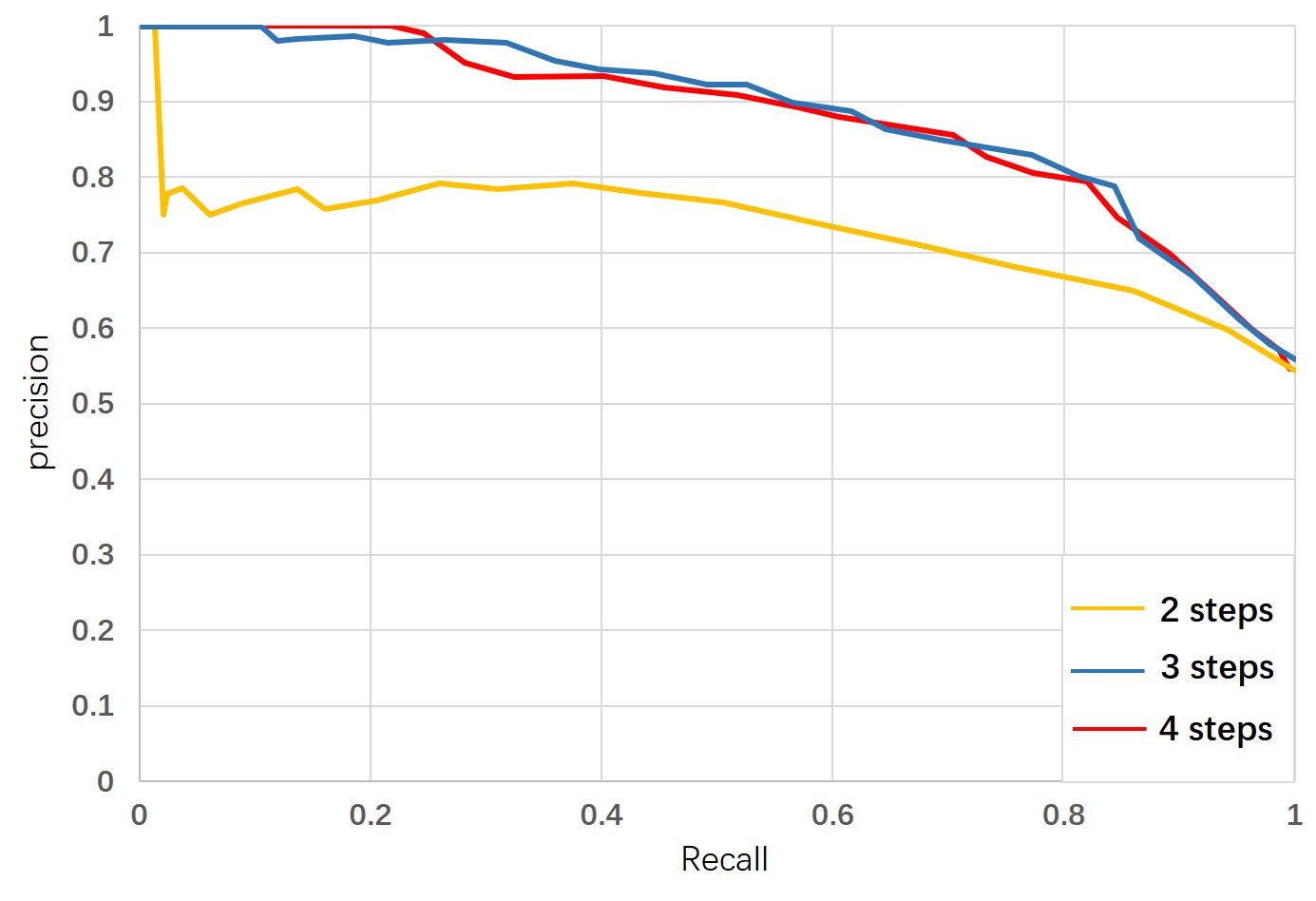}
}
\subfigure[Amount of classes]{
\centering
\includegraphics[width=2.0in]{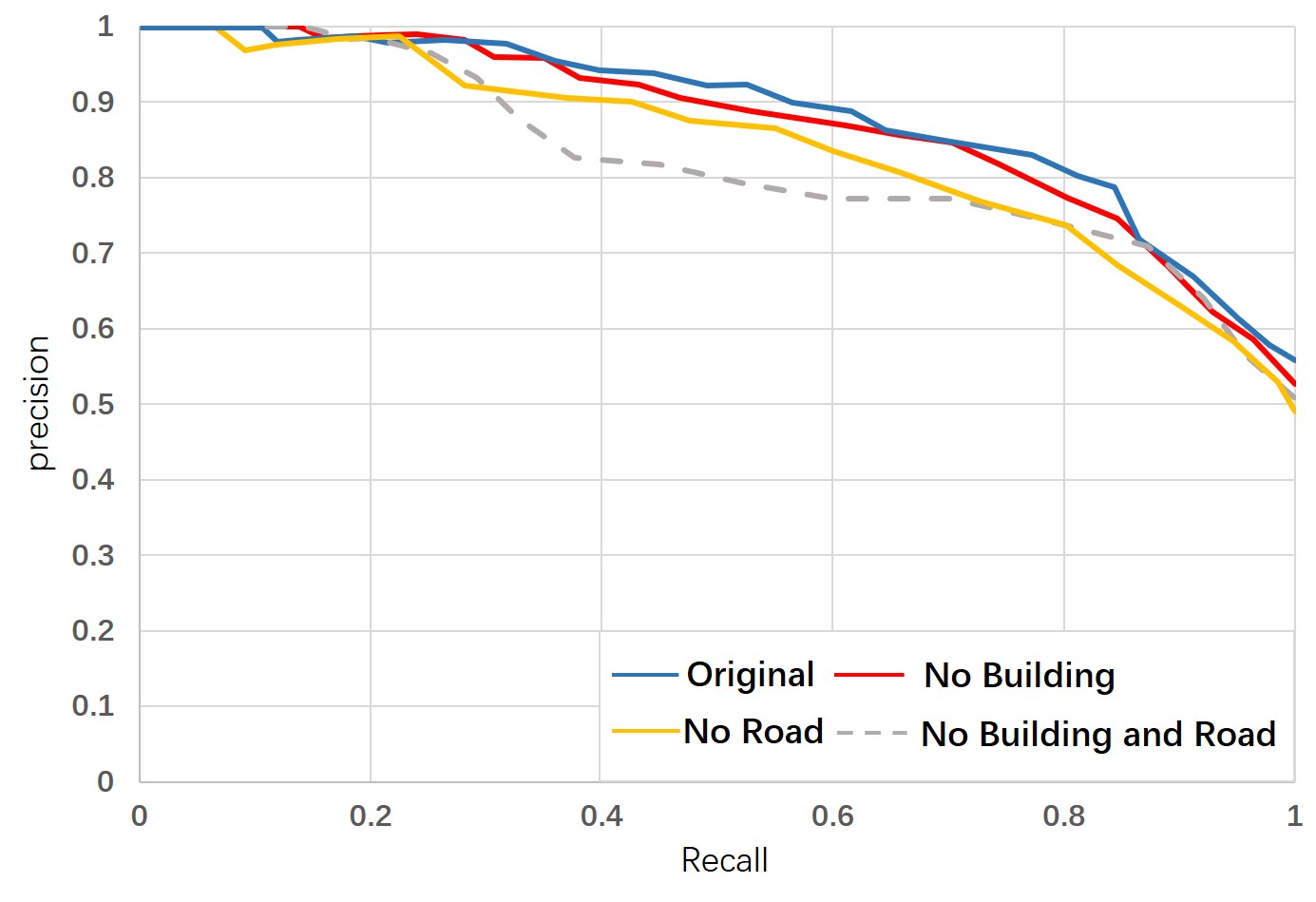}
}
\subfigure[Matching option]{
\centering
\includegraphics[width=2.0in]{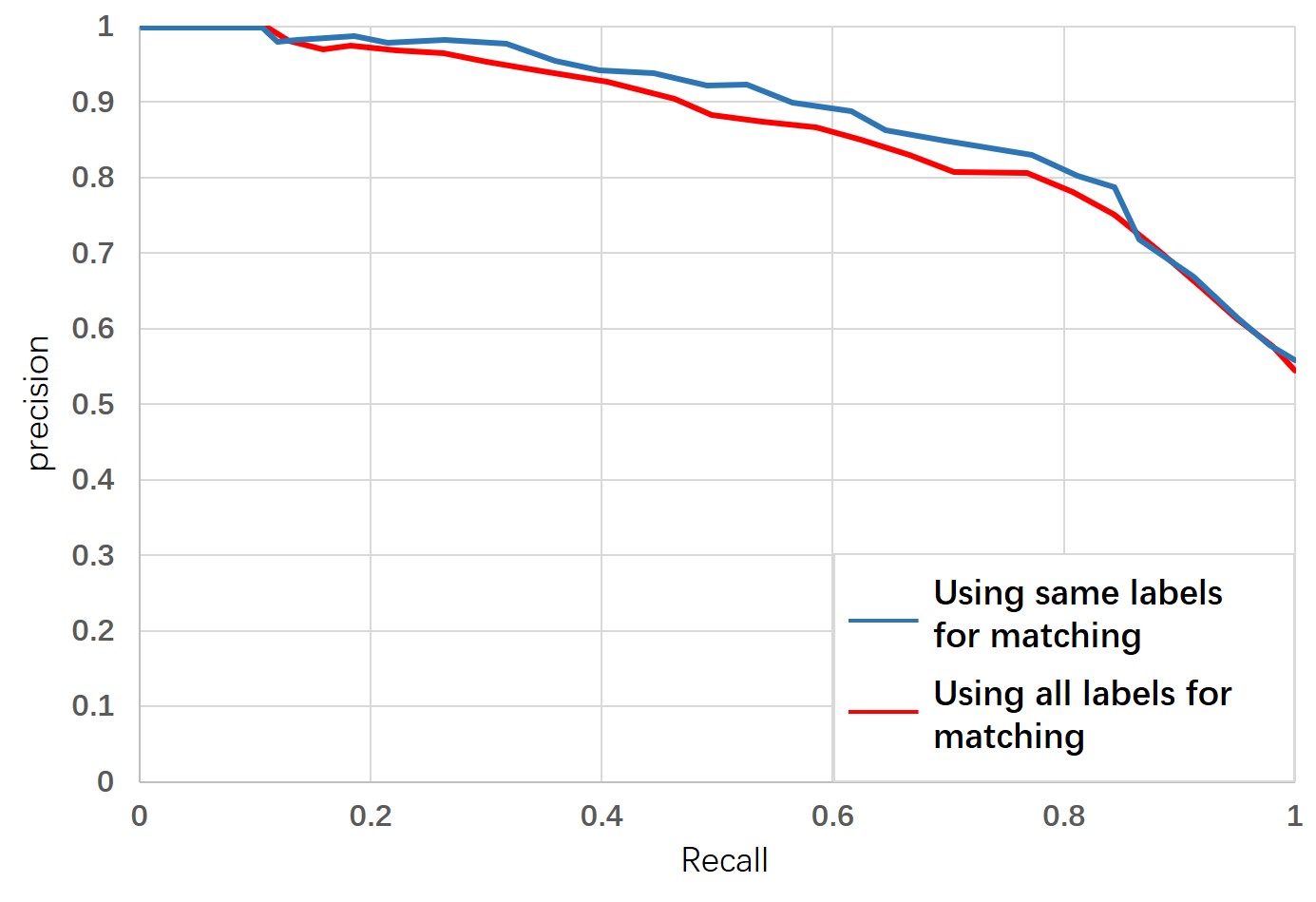}
}
\subfigure[Segmentation quality]{
\centering
\includegraphics[width=2.0in]{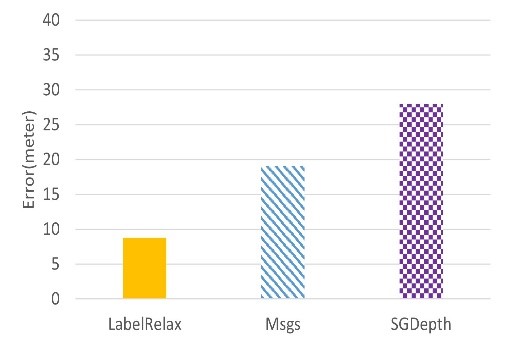}
}
\subfigure[Depth quality]{
\centering
\includegraphics[width=2.0in]{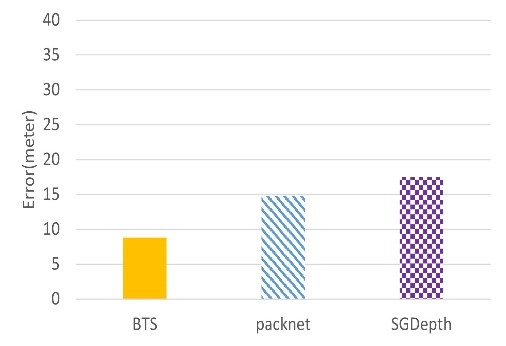}
}
 \caption{Sensitivity analysis. (a)-(d) are results on the SYNTHIA dataset, (e) and (f) are results on sequence 02 of KITTI dataset. (a) shows the translation errors with different distance threshold for setting neighbors. (b) shows localization performance of various path dimensions of descriptors. (c) shows the effect of removing different classes. (d) shows the effect of using the same label nodes or all nodes for matching. (e) show the effect of semantic maps estimated by different methods. (f) shows the effect of depth maps estimated by different methods.}
\label{fig:Sensitivity}
\vspace{-5mm}
\end{figure*}

\subsubsection{Experimental Results}

Table~\ref{KITTI_error} shows the average translation errors and their standard deviation on the KITTI datasets. As seen that BoW method demonstrates the worst performance. Our semantic histogram based descriptor archives lowest translation error on all sequences. It's noted that when the overlap is small (sequence 02 and sequence 08B), our method shows the clear advantage against others.
We can also observe that within the specific range, the longer overlap contributes to the higher localization performance. However, if the overlap length is beyond a specific range, the localization performance will decrease (sequence 19). It is due to the larger amount of disturbing objects. 

As indicated by the results, the translation error is obviously large than those on AirSim. This is because of the unavoidable error of depth and semantic prediction with deep neural networks.
Another intriguing observation is Random Walk based descriptor is worse than Neighbor Vector on the KITTI dataset, which is not consistent with the results on AirSim.
Due to the higher quantity of objects and misclassfication problem, the semantic graphs of KITTI contain higher nodes density. The higher nodes density brings more possibilities of walks. Then, the Random Walk descriptor is easier to omit the possible walks. Therefore, the performance of Random Walk descriptor is dropped significantly. By considering the unstable performance of these methods, our method performs robustly and accurately for both synthesized and real-world scenarios.
Fig.~\ref{fig:RealWorld} (b) shows the visualization of map fusion of sequence 08 after the global localization performed by our system.

\subsection{Sensitivity Analysis}

 
 In this section, we investigate the effect of different parameter settings and qualities of inputs to the performance of our method. Specifically, we conduct six ablation studies on  
connectivity threshold, path dimension, class amount (with or without different labels), matching option (all labels or same labels), segmentation and depth, respectively. The  experiments of segmentation and depth are performed on the sequence 02 of KITTI dataset and the other experiments are performed on the SYNTHIA dataset.
 

For evaluating the effect of connectivity, we use 5 meters, 10 meters, and 15 meters as the threshold of adding edges. 
The experimental results are shown in Fig.~\ref{fig:Sensitivity} (a). As seen that the performance is clearly better than others when the threshold is set to 10 meters. 
 As large threshold causes the homogenization of the descriptors and small threshold leads to many isolated nodes, thus, the performance tends to be poor for both large and small threshold values. 
Similarly, Fig.~\ref{fig:Sensitivity} (b) shows the performance of different path dimensions for descriptor extraction.  Since the lower path dimension is not able to capture sufficient surrounding information, 2 steps descriptors demonstrates the lowest localization performance. Although the performance is similar between 3 steps and 4 steps descriptors, the matching complexity of 4 steps descriptors is much higher than 3 steps descriptors, e.g. when we use 30 frames for global localization on SYNTHIA dataset, the matching time for 3 steps descriptors is around 48 ms, while the time of 4 steps descriptors is about 625 ms.

 Effect of amount of classes is compared by removing different classes including Building, Road, Building and Road. As a result, the performance of removing Road is lower than removing Building. It is also shown that removing both classes leads to further degradation of performance, as seen in Fig.~\ref{fig:Sensitivity} (c).
  Note that removing the class of large areas (for SYNTHIA dataset, the class is Building; for KITTI, the class is Vegetation) slightly lowers the performance on SYNTHIA.
On the other hand, it leads to slight performance improvement (the translation error is improved from 8.78 meters to 8.68 meters) on KITTI. Therefore, we can say that the large areas may improve or degrade the localization performance, it depends on the specific scenarios that we use.

The effect of using the same labels and all labels for graph matching is shown in Fig.~\ref{fig:Sensitivity} (d). It is observed that the performance of the same labels matching is obviously higher than the all labels matching. Note that another disadvantage of using all labels for graph matching is the significant increase of computation time. The matching time for the same label matching and all label matching is around 48 ms and 340 ms, respectively.

For ablation study on semantics, we compare LabelRelax \cite{segmentation} (higher quality), MSeg \cite{MSeg_2020_CVPR} (medium quality) and SGDepth\cite{klingner2020selfsupervised} (lower quality). For depths, we choose BTS \cite{depth} (higher quality), packnet \cite{packnet} (medium quality) and SGDepth \cite{klingner2020selfsupervised} (lower quality) for comparison.
Fig.~\ref{fig:Sensitivity} (e) and (f) show the results of different methods for semantic and depth estimation, respectively. 
The results show the more accurate semantic and depth maps we have, the more high localization performance we can gain.
Besides, we find that the localization performance is more sensitive to semantic quality than depth quality.


\section{CONCLUSIONS}
In this paper, we studied the problem of global localization for vision based multi-robot SLAM.
We argue that there are mainly two difficulties that need to be well handled. The first is the large viewpoint difference, which is ubiquitous for multi-robot systems. 
The second difficulty is the global localization needs to be performed in real-time.
These difficulties motivated us to develop a more effective and efficient method. In this paper, we proposed a semantic histogram based descriptor. Thanks to that, the graph matching is formulated as a dot-product between two descriptor sets, which can be performed in real-time. Based on the proposed descriptor, we presented a more accurate and efficient global localization system.  
The system is fairly tested on synthesized SYNTHIA, AirSim datasets as well as a real-world KITTI dataset. The experimental results show that our method outperforms others by a good margin, and it is much faster than the previous semantic based method built on the Random Walk. 



\bibliographystyle{IEEEtran}   
\bibliography{references.bib} 

\begin{thebibliography}{10}
\providecommand{\url}[1]{#1}
\csname url@rmstyle\endcsname
\providecommand{\newblock}{\relax}
\providecommand{\bibinfo}[2]{#2}
\providecommand\BIBentrySTDinterwordspacing{\spaceskip=0pt\relax}
\providecommand\BIBentryALTinterwordstretchfactor{4}
\providecommand\BIBentryALTinterwordspacing{\spaceskip=\fontdimen2\font plus
\BIBentryALTinterwordstretchfactor\fontdimen3\font minus
  \fontdimen4\font\relax}
\providecommand\BIBforeignlanguage[2]{{%
\expandafter\ifx\csname l@#1\endcsname\relax
\typeout{** WARNING: IEEEtran.bst: No hyphenation pattern has been}%
\typeout{** loaded for the language `#1'. Using the pattern for}%
\typeout{** the default language instead.}%
\else
\language=\csname l@#1\endcsname
\fi
#2}}

\bibitem{survey}
J.~Kshirsagar, S.~Shue, and J.~M. Conrad, ``A survey of implementation of
  multi-robot simultaneous localization and mapping,'' in \emph{SoutheastCon
  2018}.\hskip 1em plus 0.5em minus 0.4em\relax IEEE, 2018, pp. 1--7.

\bibitem{carlone2010rao}
L.~Carlone, M.~K. Ng, J.~Du, B.~Bona, and M.~Indri, ``Rao-blackwellized
  particle filters multi robot slam with unknown initial correspondences and
  limited communication,'' in \emph{2010 IEEE International Conference on
  Robotics and Automation}.\hskip 1em plus 0.5em minus 0.4em\relax IEEE, 2010,
  pp. 243--249.

\bibitem{X-view}
A.~Gawel, C.~D. Don, R.~Siegwart, J.~Nieto, and C.~Cadena, ``X-view:
  Graph-based semantic multi-view localization,'' \emph{IEEE Robotics and
  Automation Letters}, vol.~3, no.~3, p. 1687–1694, 2018.

\bibitem{lee2012survey}
H.-C. Lee, S.-H. Lee, T.-S. Lee, D.-J. Kim, and B.-H. Lee, ``A survey of map
  merging techniques for cooperative-slam,'' in \emph{2012 9th International
  Conference on Ubiquitous Robots and Ambient Intelligence (URAI)}.\hskip 1em
  plus 0.5em minus 0.4em\relax IEEE, 2012, pp. 285--287.

\bibitem{bow}
D.~Galvez-López and J.~D. Tardos, ``Bags of binary words for fast place
  recognition in image sequences,'' \emph{IEEE Transactions on Robotics},
  vol.~28, no.~5, p. 1188–1197, 2012.

\bibitem{FAB}
M.~Cummins and P.~Newman, ``Fab-map: Probabilistic localization and mapping in
  the space of appearance,'' \emph{The International Journal of Robotics
  Research}, vol.~27, no.~6, p. 647–665, 2008.

\bibitem{CCM}
P.~Schmuck and M.~Chli, ``Multi-uav collaborative monocular slam,'' \emph{2017
  IEEE International Conference on Robotics and Automation (ICRA)}, 2017.

\bibitem{dis-SLAM}
X.~Chen, H.~Lu, J.~Xiao, and H.~Zhang, ``Distributed monocular multi-robot
  slam,'' \emph{2018 IEEE 8th Annual International Conference on CYBER
  Technology in Automation, Control, and Intelligent Systems (CYBER)}, 2018.

\bibitem{SIFT}
D.~G. Lowe, ``Distinctive image features from scale-invariant keypoints,''
  \emph{International Journal of Computer Vision}, vol.~60, no.~2, p. 91–110,
  2004.

\bibitem{SURF}
H.~Bay, A.~Ess, T.~Tuytelaars, and L.~V. Gool, ``Speeded-up robust features
  (surf),'' \emph{Computer Vision and Image Understanding}, vol. 110, no.~3, p.
  346–359, 2008.

\bibitem{ORB-detector}
E.~Rublee, V.~Rabaud, K.~Konolige, and G.~Bradski, ``Orb: An efficient
  alternative to sift or surf,'' \emph{2011 International Conference on
  Computer Vision}, 2011.

\bibitem{FAST}
E.~Rosten and T.~Drummond, ``Machine learning for high-speed corner
  detection,'' \emph{Computer Vision – ECCV 2006 Lecture Notes in Computer
  Science}, p. 430–443, 2006.

\bibitem{indoor1}
Y.~Liu, Y.~Petillot, D.~Lane, and S.~Wang, ``Global localization with
  object-level semantics and topology,'' \emph{2019 International Conference on
  Robotics and Automation (ICRA)}, 2019.

\bibitem{indoor2}
J.~Li, D.~Meger, and G.~Dudek, ``Semantic mapping for view-invariant
  relocalization,'' \emph{2019 International Conference on Robotics and
  Automation (ICRA)}, 2019.

\bibitem{ConvNet}
N.~Suenderhauf, S.~Shirazi, A.~Jacobson, F.~Dayoub, E.~Pepperell, B.~Upcroft,
  and M.~Milford, ``Place recognition with convnet landmarks: Viewpoint-robust,
  condition-robust, training-free,'' \emph{Robotics: Science and Systems XI},
  2015.

\bibitem{DeepLearning}
Z.~Chen, A.~Jacobson, N.~Sunderhauf, B.~Upcroft, L.~Liu, C.~Shen, I.~Reid, and
  M.~Milford, ``Deep learning features at scale for visual place recognition,''
  \emph{2017 IEEE International Conference on Robotics and Automation (ICRA)},
  2017.

\bibitem{NetVLAD}
R.~Arandjelovic, P.~Gronat, A.~Torii, T.~Pajdla, and J.~Sivic, ``Netvlad: Cnn
  architecture for weakly supervised place recognition,'' \emph{2016 IEEE
  Conference on Computer Vision and Pattern Recognition (CVPR)}, 2016.

\bibitem{lost}
S.~Garg, N.~Suenderhauf, and M.~Milford, ``Lost? appearance-invariant place
  recognition for opposite viewpoints using visual semantics,'' \emph{Robotics:
  Science and Systems XIV}, 2018.

\bibitem{loolnoDeeper}
S.~Garg, M.~B. V, T.~Dharmasiri, S.~Hausler, N.~Suenderhauf, S.~Kumar,
  T.~Drummond, and M.~Milford, ``Look no deeper: Recognizing places from
  opposing viewpoints under varying scene appearance using single-view depth
  estimation,'' \emph{2019 International Conference on Robotics and Automation
  (ICRA)}, 2019.

\bibitem{ORB-SLAM3}
C.~Campos, R.~Elvira, J.~J.~G. Rodr{\'\i}guez, J.~M. Montiel, and J.~D.
  Tard{\'o}s, ``Orb-slam3: An accurate open-source library for visual,
  visual-inertial and multi-map slam,'' \emph{arXiv preprint arXiv:2007.11898},
  2020.

\bibitem{angelina2018pointnetvlad}
M.~Angelina~Uy and G.~Hee~Lee, ``Pointnetvlad: Deep point cloud based retrieval
  for large-scale place recognition,'' in \emph{Proceedings of the IEEE
  Conference on Computer Vision and Pattern Recognition}, 2018, pp. 4470--4479.

\bibitem{graph1}
E.~Stumm, C.~Mei, S.~Lacroix, and M.~Chli, ``Location graphs for visual place
  recognition,'' \emph{2015 IEEE International Conference on Robotics and
  Automation (ICRA)}, 2015.

\bibitem{graph2}
E.~Stumm, C.~Mei, S.~Lacroix, J.~Nieto, M.~Hutter, and R.~Siegwart, ``Robust
  visual place recognition with graph kernels,'' \emph{2016 IEEE Conference on
  Computer Vision and Pattern Recognition (CVPR)}, 2016.

\bibitem{graphseg1}
R.~Finman, L.~Paul, and J.~L. Leonard, ``Toward object-based place recognition
  in dense rgb-d maps,'' \emph{ICRA Workshop on Visual Place Recognition in
  Changing Environments}, 2015.

\bibitem{seedfilling}
M.~C. Codrea and O.~S. Nevalainen, ``Note: An algorithm for contour-based
  region filling,'' \emph{Computers \& Graphics}, vol.~29, no.~3, pp. 441--450,
  2005.

\bibitem{ICP}
P.~J. Besl and N.~D. McKay, ``Method for registration of 3-d shapes,'' in
  \emph{Sensor fusion IV: control paradigms and data structures}, vol.
  1611.\hskip 1em plus 0.5em minus 0.4em\relax International Society for Optics
  and Photonics, 1992, pp. 586--606.

\bibitem{RANSAC}
M.~A. Fischler and R.~C. Bolles, ``Random sample consensus: a paradigm for
  model fitting with applications to image analysis and automated
  cartography,'' \emph{Communications of the ACM}, vol.~24, no.~6, pp.
  381--395, 1981.

\bibitem{SYNTHIA}
G.~Ros, L.~Sellart, J.~Materzynska, D.~Vazquez, and A.~M. Lopez, ``The synthia
  dataset: A large collection of synthetic images for semantic segmentation of
  urban scenes,'' \emph{2016 IEEE Conference on Computer Vision and Pattern
  Recognition (CVPR)}, 2016.

\bibitem{AirSim}
S.~Shah, D.~Dey, C.~Lovett, and A.~Kapoor, ``Airsim: High-fidelity visual and
  physical simulation for autonomous vehicles,'' \emph{Field and Service
  Robotics Springer Proceedings in Advanced Robotics}, p. 621–635, 2017.

\bibitem{KITTI}
A.~Geiger, P.~Lenz, and R.~Urtasun, ``Are we ready for autonomous driving? the
  kitti vision benchmark suite,'' \emph{2012 IEEE Conference on Computer Vision
  and Pattern Recognition(CVPR)}, 2012.

\bibitem{depth}
J.~H. Lee, M.-K. Han, D.~W. Ko, and I.~H. Suh, ``From big to small: Multi-scale
  local planar guidance for monocular depth estimation,'' \emph{arXiv preprint
  arXiv:1907.10326}, 2019.

\bibitem{segmentation}
Y.~Zhu, K.~Sapra, F.~A. Reda, K.~J. Shih, S.~Newsam, A.~Tao, and B.~Catanzaro,
  ``Improving semantic segmentation via video propagation and label
  relaxation,'' \emph{2019 IEEE/CVF Conference on Computer Vision and Pattern
  Recognition (CVPR)}, 2019.

\bibitem{MSeg_2020_CVPR}
J.~Lambert, Z.~Liu, O.~Sener, J.~Hays, and V.~Koltun, ``{MSeg}: A composite
  dataset for multi- domain semantic segmentation,'' in \emph{Computer Vision
  and Pattern Recognition (CVPR)}, 2020.

\bibitem{klingner2020selfsupervised}
M.~Klingner, J.-A. Termöhlen, J.~Mikolajczyk, and T.~Fingscheidt,
  ``{Self-Supervised Monocular Depth Estimation: Solving the Dynamic Object
  Problem by Semantic Guidance},'' in \emph{ECCV}, 2020.

\bibitem{packnet}
V.~Guizilini, R.~Ambrus, S.~Pillai, A.~Raventos, and A.~Gaidon, ``3d packing
  for self-supervised monocular depth estimation,'' in \emph{IEEE Conference on
  Computer Vision and Pattern Recognition (CVPR)}, 2020.

\end{thebibliography}

\ifCLASSOPTIONcaptionsoff
  \newpage
\fi



%




\end{document}